\documentclass{article}
\usepackage{
spconf,
amsmath,
graphicx,
url,
booktabs,
enumitem,
}

\title{Benchmarking Cognitive Domains for LLMs: Insights from Taiwanese Hakka Culture}

\name{Chen-Chi Chang$^1$, Ching-Yuan Chen$^1$, Hung-Shin Lee$^3$, and Chih-Cheng Lee$^2$}
\address{$^1$Dept. Culture Creativity and Digital Marketing, National United University, Taiwan\\
$^2$Dept. Information Management, National United University, Taiwan\\
$^3$United Link Co., Ltd., Taiwan}

\begin{document}
\maketitle
\begin{abstract}
This study introduces a comprehensive benchmark designed to evaluate the performance of large language models (LLMs) in understanding and processing cultural knowledge, with a specific focus on Hakka culture as a case study. Leveraging Bloom's Taxonomy, the study develops a multi-dimensional framework that systematically assesses LLMs across six cognitive domains: Remembering, Understanding, Applying, Analyzing, Evaluating, and Creating. This benchmark extends beyond traditional single-dimensional evaluations by providing a deeper analysis of LLMs' abilities to handle culturally specific content, ranging from basic recall of facts to higher-order cognitive tasks such as creative synthesis. Additionally, the study integrates Retrieval-Augmented Generation (RAG) technology to address the challenges of minority cultural knowledge representation in LLMs, demonstrating how RAG enhances the models' performance by dynamically incorporating relevant external information. The results highlight the effectiveness of RAG in improving accuracy across all cognitive domains, particularly in tasks requiring precise retrieval and application of cultural knowledge. However, the findings also reveal the limitations of RAG in creative tasks, underscoring the need for further optimization. This benchmark provides a robust tool for evaluating and comparing LLMs in culturally diverse contexts, offering valuable insights for future research and development in AI-driven cultural knowledge preservation and dissemination.
\end{abstract}
\section{Introduction}
\label{sec:intro}

Minority cultures, particularly those with low-resource languages, are integral to global heritage but face unprecedented challenges in the age of AI, particularly from large language models (LLMs). While LLMs offer potential for cultural preservation through their ability to process and generate cultural content, their reliance on vast datasets and algorithms may perpetuate biases against minority groups, leading to misrepresentation. This is exacerbated by the phenomenon of ``AI hallucinations'', where LLMs generate false or misleading information, potentially distorting the authentic history and values of minority cultures \cite{maleki2024}. Consequently, preserving and accurately disseminating minority culture within the evolving AI landscape is paramount.

LLMs, trained on vast datasets, offer unprecedented opportunities for understanding and disseminating cultural knowledge. They can process, generate, and translate text, facilitating cross-cultural communication and knowledge dissemination, especially for minority groups with limited resources. LLMs can preserve and make these cultures accessible to wider audiences while breaking down language barriers through multilingual content generation \cite{chari2022}. However, their effectiveness hinges on accuracy and impartiality in their design and training. If not carefully managed, LLMs could exacerbate cultural misunderstandings. Therefore, utilizing LLMs for accurate cultural knowledge dissemination is crucial. Notably, incorporating minority data in AI training benefits both minority and majority groups, significantly reducing test errors for both \cite{chari2022}.

This study examines the potential of AI and LLMs in preserving and disseminating the Hakka language and culture, classified as low-resource due to limited linguistic materials. Using Hakka culture as a case study, the research proposes a question-and-answer test dataset focused on Hakka culture, integrating Bloom's Taxonomy and Retrieval-Augmented Generation (RAG) technology.

Bloom's Taxonomy, a hierarchical model for assessing cognitive skills \cite{bloom1956,furst1981,seddon1978}, enables a comprehensive evaluation of LLMs' abilities across various cognitive levels related to Hakka cultural knowledge. RAG technology enhances performance by retrieving information from external databases, improving the accuracy and reliability of responses \cite{lewis2020}. This approach identifies the strengths and weaknesses of LLMs in understanding Hakka culture, providing data for future model optimization and broader cultural knowledge dissemination.

The study investigates whether LLMs can fully comprehend and apply Hakka cultural knowledge, crucial for accurate cultural preservation and dissemination. Employing Bloom's Taxonomy, the research analyzes LLMs' performance across cognitive levels, from memory to creation, examining their precision and depth in processing Hakka cultural content. Additionally, the study explores if RAG technology can enhance LLMs' performance in cultural knowledge tasks, ultimately determining if these technologies can effectively support the accurate dissemination of cultural knowledge.

\section{Literature Review}
\subsection{Cognitive Domain}

Bloom's Taxonomy, developed in 1956 \cite{bloom1956}, provides a hierarchical framework for categorizing cognitive learning objectives. It has become an indispensable tool in education for designing curricula and assessing student learning. Its six levels (Remembering, Understanding, Applying, Analyzing, Evaluating, and Creating) encompass cognitive activities from basic recall to higher-order thinking and creation.

While initially designed for traditional education, Bloom's Taxonomy has seamlessly adapted to the demands of digital learning environments \cite{amin2020}. It serves as a valuable theoretical foundation for knowledge assessment in digital learning, integrating traditional cognitive levels with modern technological tools and activities. By applying this structured framework, the study comprehensively evaluates how well LLMs grasp and utilize Hakka cultural knowledge across various cognitive levels. This provides valuable insights for cultural preservation, education, and AI development in culturally-sensitive contexts \cite{jiang2024, poornima2024, spanos2024, wang2024}.

\subsection{Retrieval-Augmented Generation (RAG)}

LLMs, while powerful, face challenges like hallucinations and outdated information. RAG addresses these limitations by integrating LLMs with external knowledge bases, significantly enhancing their accuracy and reliability, particularly in domain-specific tasks \cite{gao2023}. Its strength lies in two-step process: 1) Retrieval: Based on the input query, the model retrieves relevant information from a predefined knowledge base (e.g., Wikipedia) \cite{yu2022}; 2) Generation: The retrieved information is integrated with the original input to generate a comprehensive and accurate response.

This approach, combining internal LLM knowledge with external, dynamic data sources, allows for continuous knowledge updates and seamless integration of specialized information. This is particularly valuable in domains like medicine or cultural studies, where accuracy and up-to-date information are critical. By enhancing the retrieval and generation capabilities of LLMs, RAG significantly improves their performance in domain-specific tasks, making them more reliable and valuable tools for various professional applications.

\subsection{Application of LLMs in Cultural Knowledge Assessment}

LLMs offer both challenges and opportunities when applied to cultural knowledge assessment, particularly for minority cultures. Their reliance on vast datasets, often dominated by mainstream cultural knowledge, can lead to inaccuracies when handling the nuances of under-represented cultures.

Traditional evaluation methods, like MMLU \cite{hendrycks2020}, SQuAD \cite{rondeau2018}, and XSum \cite{narayan2018}, while valuable for assessing general knowledge and language comprehension, are limited in their ability to accurately evaluate understanding of minority cultures due to the lack of representative data in their training sets. However, LLMs, especially when enhanced with techniques like RAG, hold immense potential for promoting and preserving minority cultural knowledge:
\begin{enumerate}[noitemsep,leftmargin=*]
\item \textbf{Addressing Data Scarcity}: RAG enables LLMs to access and learn from specialized cultural resources, such as oral histories and local literature, that are often absent in standard training data.
\item \textbf{Facilitating Accurate Representation}: By incorporating diverse cultural perspectives, LLMs can contribute to more nuanced and respectful representations of minority cultures.
\item \textbf{Promoting Global Dissemination}: LLMs can make minority cultural knowledge more accessible to wider audiences, fostering cross-cultural understanding and appreciation.
\end{enumerate}
To fully leverage these opportunities, it is crucial to develop specialized test datasets and evaluation methods tailored to specific minority cultures; prioritize the representation of diverse cultural knowledge in LLM training data; continue refining techniques like RAG to enhance the accuracy and sensitivity of LLMs in cultural contexts.

By addressing these challenges, we can harness the power of LLMs to become valuable tools for preserving, understanding, and celebrating the richness of minority cultures in the digital age.

\section{Research Methodology}

This study develops a framework to evaluate how well AI LLMs comprehend different cognitive domain levels, adapting Bloom's Taxonomy from education to assess LLM training outcomes. This framework is applied to create a Hakka cultural knowledge test set, focusing on this low-resource language and culture.

To ensure comprehensive coverage, the study draws on diverse sources like historical documents, academic research, and oral histories, encompassing aspects like language, architecture, cuisine, and festivals. This rich dataset forms the basis for a multi-layered question set designed using Bloom’s Taxonomy's six levels.

Questions range from basic recall (e.g., ``What are the main Hakka settlements?'') to higher-order creative application (e.g., ``Design a modern project promoting Hakka culture''). This approach evaluates not only the LLM's grasp of fundamental cultural knowledge but also its reasoning and creative abilities within a cultural context, providing a robust framework for assessing overall performance.

\begin{figure}[t]
\centering
\includegraphics[width=1.0\linewidth]{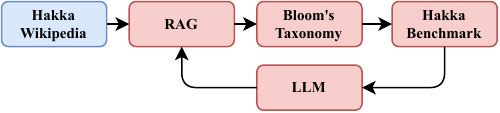}
\vspace{-15pt}
\caption{Our proposed research procedure.}
\label{fig:procedure}
\vspace{-10pt}
\end{figure}

\subsection{Research Design}


This study utilizes a framework integrating Bloom's Taxonomy and Retrieval-Augmented Generation (RAG) technology to comprehensively evaluate how well LLMs understand and apply Hakka cultural knowledge (see Fig. \ref{fig:procedure}).

Bloom's Taxonomy's six cognitive levels, from memory to creation, are used to design a multi-layered question set. This set assesses diverse cognitive abilities, from basic recall to complex creative thinking, ensuring a thorough evaluation of the LLM's depth and breadth of understanding. The questions are designed to be diverse, linguistically accurate, and culturally relevant, encompassing Hakka language, customs, history, architecture, and more.

Furthermore, RAG technology integrates external database information into the LLM's response process, enhancing the precision and reliability of answers. This framework not only evaluates the LLM's cultural knowledge base but also its ability to retrieve and generate knowledge in real-world scenarios, offering valuable insights for optimizing these models and their applications.

\subsection{Cognitive Domain Benchmark Construction}

This study constructs a benchmark dataset to evaluate how well LLMs comprehend and apply Hakka cultural knowledge across Bloom's Taxonomy's six cognitive levels: Remembering, Understanding, Applying, Analyzing, Evaluating, and Creating \cite{sarkis2001,talluri2001}.
The dataset draws primarily from the Hakka Culture Encyclopedia, focusing on the Hakka community in Taiwan's Miaoli region. This encyclopedia, with its 2,029 entries across 22 categories, covers diverse cultural aspects. The Taiwan Ministry of Education's Hakka Knowledge Base supplements this primary source.
The dataset comprises 36,522 questions, with three questions for each of the six Bloom's Taxonomy categories generated from the 2,029 encyclopedia entries. RAG techniques are integrated to enhance the LLMs' understanding and processing of the culturally rich questions, improving the accuracy and relevance of their responses. Each question type corresponds to a Bloom's Taxonomy level:
\begin{enumerate}[noitemsep,leftmargin=*]
\item \textbf{Remembering}: Recall of facts (e.g., ``What are Miaoli County's main Hakka settlements?'')
\item \textbf{Understanding}: Comprehension of concepts (e.g., ``What distinguishes Hakka culinary practices?'')
\item \textbf{Applying}: Applying knowledge (e.g., ``How can Hakka clothing styles be used in modern fashion?'')
\item \textbf{Analyzing}: Examining and comparing (e.g., ``Compare Hakka opera with other Han Chinese theater forms.'')
\item \textbf{Evaluating}: Critical assessment (e.g., ``Evaluate the historical evolution of Hakka social status in Taiwan.'')
\item \textbf{Creating}: Generating innovative ideas (e.g., ``Design an art exhibition showcasing Hakka heritage.'')
\end{enumerate}
This comprehensive dataset allows for a thorough evaluation of the LLM's ability to understand, process, and apply Hakka cultural knowledge within a robust cognitive framework.

The benchmark dataset construction process ensures a comprehensive evaluation of LLMs' ability to understand and engage with Hakka cultural knowledge, as well as the effectiveness of RAG techniques in enhancing their cultural cognition. The process involves four key steps:
\begin{enumerate}[noitemsep,leftmargin=*]
\item \textbf{Compilation of Hakka Cultural Texts}: Gathering extensive textual resources from the Hakka Culture Encyclopedia and the Ministry of Education's Hakka Knowledge Base.
\item \textbf{Development of a Specialized QA Knowledge Base}: Utilizing the compiled texts, RAG techniques, and the ChatGPT language model to create a domain-specific question-and-answer repository for each of the 2,029 encyclopedia entries.
\item \textbf{Design of Cognitive Domain Questions}: For each of the 1,693 selected entries, multiple-choice questions and corresponding answers are formulated based on Bloom's Taxonomy's six levels, resulting in the Hakka Cultural Knowledge Cognitive Domain Dataset containing 10,158 multiple-choice questions.
\item \textbf{Evaluation of Language Models}: Utilizing the constructed dataset to systematically assess and benchmark various state-of-the-art LLMs on their understanding and cognitive processing of Hakka cultural knowledge across all defined cognitive levels. 
\end{enumerate}
This rigorous process ensures the dataset's effectiveness in evaluating LLMs' cultural knowledge comprehension and the impact of RAG techniques on their performance.

\subsection{Model Training and Testing}

For model training and testing, this study selected several advanced LLMs: GPT-4o, Gemini 1.5 Pro, Claude 3.5 Sonnet, Llama 3.1 (70b version), and LLaMA 3.1 with RAG technology. RAG integration was applied only to LLaMA 3.1, the lowest performer among the selected LLMs, to investigate if RAG could improve its understanding and accuracy regarding Hakka cultural knowledge.

During training, each model received preliminary fine-tuning with foundational Hakka cultural knowledge. Subsequently, RAG technology was introduced, allowing models to access and utilize external resources like the Hakka Culture Encyclopedia during question answering, leading to more accurate and comprehensive responses. This was particularly effective in addressing knowledge gaps related to less common or highly specialized cultural aspects, ultimately enhancing the model's reasoning and generative capabilities.

Model evaluation involved the previously constructed dataset structured according to Bloom's Taxonomy's six levels. Performance metrics, including accuracy rate, were collected for each model across all cognitive levels. This provided a detailed analysis of each model's strengths and weaknesses in understanding and applying Hakka cultural knowledge, offering valuable insights for future model optimization and enhancing the effectiveness of RAG technology within this cultural domain.

\begin{table}[t]
\small
\caption{Accuracy (\%) with respect to various LLMs, where \#Q and \#CA denote the numbers of questions and correct answers, respectively.}
\vspace{5pt}
\label{tab:result}
\centering
\setlength{\tabcolsep}{5pt}
\begin{tabular}{lcccc}
\toprule
\bf Model & \bf Category & \bf \#CA & \bf \#Q & \bf Acc. \\
\hline
GPT-4o & Remembering & 1,230 & 1,693 & 72.65 \\
Gemini 1.5 Pro & Remembering & 1,183 & 1,693 & 69.88 \\
Claude 3.5 Sonnet & Remembering & 1,232 & 1,693 & 72.77 \\
Llama 3.1 & Remembering & 1,043 & 1,693 & 61.61 \\
Llama 3.1 with RAG & Remembering & 1,542 & 1,693 & \bf 91.08 \\
\hline
GPT-4o & Understanding & 1,442 & 1,693 & 85.17 \\
Gemini 1.5 Pro & Understanding & 1,386 & 1,693 & 81.87 \\
Claude 3.5 Sonnet & Understanding & 1,487 & 1,693 & 87.83 \\
Llama 3.1 & Understanding & 1344 & 1,693 & 79.39 \\
Llama 3.1 with RAG & Understanding & 1,601 & 1,693 & \bf 94.57 \\
\hline
GPT-4o & Applying & 1,364 & 1,693 & 80.57 \\
Gemini 1.5 Pro & Applying & 1,315 & 1,693 & 77.67 \\
Claude 3.5 Sonnet & Applying & 1,394 & 1,693 & 82.34 \\
Llama 3.1 & Applying & 1,270 & 1,693 & 75.01 \\
Llama 3.1 with RAG & Applying & 1,542 & 1,693 & \bf 91.08 \\
\hline
GPT-4o & Analyzing & 1,370 & 1,693 & 80.92 \\
Gemini 1.5 Pro & Analyzing & 1,354 & 1,693 & 79.98 \\
Claude 3.5 Sonnet & Analyzing & 1,418 & 1,693 & 83.76 \\
Llama 3.1 & Analyzing & 1,273 & 1,693 & 75.19 \\
Llama 3.1 with RAG & Analyzing & 1,528 & 1,693 & \bf 90.25 \\
\hline
GPT-4o & Evaluating & 1,445 & 1,693 & 85.35 \\
Gemini 1.5 Pro & Evaluating & 1,410 & 1,693 & 83.28 \\
Claude 3.5 Sonnet & Evaluating & 1,479 & 1,693 & 87.36 \\
Llama 3.1 & Evaluating & 1,423 & 1,693 & 84.05 \\
Llama 3.1 with RAG & Evaluating & 1,516 & 1,693 & \bf 89.55 \\
\hline
GPT-4o & Creating & 1,488 & 1,693 & \bf 87.89 \\
Gemini 1.5 Pro & Creating & 1,402 & 1,693 & 82.81 \\
Claude 3.5 Sonnet & Creating & 1,455 & 1,693 & 85.94 \\
Llama 3.1 & Creating & 1,364 & 1,693 & 80.57 \\
Llama 3.1 with RAG & Creating & 1,434 & 1,693 & 84.70 \\
\hline
GPT-4o & Overall & 8,339 & 10,158 & 82.09 \\
Gemini 1.5 Pro & Overall & 8,050 & 10,158 & 79.25 \\
Claude 3.5 Sonnet & Overall & 8,465 & 10,158 & 83.33 \\
Llama 3.1 & Overall & 7,717 & 10,158 & 75.97 \\
Llama 3.1 with RAG & Overall & 9,163 & 10,158 & \bf 90.20 \\
\bottomrule
\end{tabular}
\vspace{-10pt}
\end{table}

\subsection{Data Analysis}


This study conducted a thorough performance analysis of the selected LLMs, focusing on their understanding and application of Hakka cultural knowledge, especially when integrated with RAG. A quantitative approach was employed, using automated scoring against standard answers for the multiple-choice questions to assess model accuracy. This method allowed for a precise evaluation of how well each LLM comprehended and responded to the Hakka cultural knowledge questions. The analysis results, presented as comparative performance metrics across the cognitive domains defined by Bloom's Taxonomy, are available in Table \ref{tab:result}. This comparative analysis provides insights into the strengths and weaknesses of each LLM and the impact of RAG technology on their performance in understanding and applying Hakka cultural knowledge.

\section{Discussion}
\subsection{Overall Performance Across Cognitive Domains}


Analysis of the four LLMs (GPT-4o, Gemini 1.5 Pro, Claude 3.5 Sonnet, and Llama 3.1) reveals varying degrees of accuracy in understanding and applying Hakka cultural knowledge across Bloom's Taxonomy's cognitive domains. Notably, integrating RAG technology significantly improved LLaMA 3.1's performance across all domains. For instance, in the Remembering domain, accuracy increased from 61.61\% without RAG to 91.08\% with RAG. Similar improvements were observed across all other domains, demonstrating RAG's effectiveness in enhancing the model's ability to process and generate accurate responses related to Hakka culture.

Overall, LLaMA 3.1 with RAG achieved the highest accuracy (90.20\%), followed by Claude 3.5 Sonnet (83.33\%), GPT-4o (82.09\%), and Gemini 1.5 Pro (79.25\%). These findings highlight the potential of RAG in mitigating the limitations of baseline LLMs in handling culturally specific knowledge, particularly for tasks requiring higher-order cognitive skills. This has significant implications for preserving and understanding minority cultures like the Hakka.

\subsection{Comparative Analysis by Domain}

Table \ref{tab:result} offers a comparative overview of GPT-4o, Claude 3.5 Sonnet, LLaMA 3.1, and Gemini 1.5 Pro's performance across the six cognitive domains. The results reveal distinct strengths and weaknesses in handling Hakka cultural knowledge. LLaMA 3.1 with RAG consistently outperforms other models across all domains, particularly excelling in tasks demanding higher-order cognitive skills. For instance, in the Remembering domain, it achieved 91.08\% accuracy, significantly surpassing others. Similarly, it led in Understanding with 94.57\% accuracy, indicating a strong grasp of cultural concepts. This trend continues in Applying, where LLaMA 3.1 with RAG achieved 91.08\% accuracy, showcasing its proficiency in practical application of knowledge. In more complex domains like Analyzing, Evaluating, and Creating, LLaMA 3.1 with RAG maintained its lead, demonstrating the effectiveness of RAG in enhancing accuracy and effectiveness for tasks requiring higher-order cognitive skills.

These findings underscore the importance of RAG in improving LLM performance in specialized cultural knowledge domains. LLaMA 3.1 with RAG's consistent superiority across all cognitive domains, especially those requiring higher-order skills, highlights its potential for broader application in cultural preservation and understanding.

\subsection{Detailed Analysis of in the Creating Domain}

The Creating domain, the most complex in Bloom's Taxonomy, offers key insights into the LLMs' capacity for generating innovative, culturally-rooted content.

GPT-4o excels in this domain, achieving 87.89\% accuracy, demonstrating its strength in synthesizing information and generating novel, culturally-aligned ideas. Claude 3.5 Sonnet follows closely (85.94\%), also showing strong creative abilities. While LLaMA 3.1 with RAG leads in other domains, it scores comparatively lower (84.70\%) in Creating. This suggests that while RAG enhances accuracy in retrieving and analyzing existing information, it might be less advantageous than inherent model capabilities when generating entirely new ideas. Gemini 1.5 Pro records the lowest accuracy (82.81\%) in this domain, indicating potential limitations in creatively applying cultural knowledge or generating original content within the specific cultural context.

These observations highlight the nuanced differences in how each model approaches creative generation. While GPT-4o and Claude 3.5 Sonnet excel in innovation, LLaMA 3.1 with RAG might be better suited for tasks focused on existing knowledge. Gemini 1.5 Pro shows a relative weakness in creative generation within this cultural context, suggesting areas for improvement. These insights are vital for understanding each model's strengths and weaknesses in culturally-sensitive creative tasks.

\section{Conclusion}
\subsection{Understanding LLM Performance in Cultural Knowledge Through Bloom's Taxonomy Framework}

This study pioneers a comprehensive evaluation of LLMs using Bloom's Taxonomy, moving beyond traditional, one-dimensional assessments that primarily focus on basic information recall. By employing Bloom's six cognitive levels—Remembering, Understanding, Applying, Analyzing, Evaluating, and Creating—this research systematically analyzes LLM performance across multiple cognitive processes.

This multi-level framework allows for a deeper understanding of LLMs' capabilities. It not only assesses their grasp of basic knowledge but also evaluates their ability to apply, analyze, evaluate, and even creatively utilize cultural knowledge, aspects often overlooked in previous research. For instance, the Analyzing level examines an LLM's ability to deconstruct and compare cultural phenomena, crucial for understanding nuanced cultural contexts. The Evaluating level assesses critical thinking and judgment of cultural knowledge, while the Creating level challenges the model's capacity for innovation based on cultural understanding.

This approach, therefore, offers a more realistic picture of how LLMs might handle cultural knowledge in real-world applications. It moves beyond simply measuring accuracy in basic recall and delves into higher-order cognitive processes, providing a more thorough and insightful evaluation.

This comprehensive analysis, enabled by the Bloom's Taxonomy framework, marks a significant advancement from previous superficial evaluations. It provides a more scientific and multi-faceted understanding of LLMs' strengths and weaknesses in processing cultural knowledge, paving the way for more targeted model optimization and informing future research in multicultural contexts.

\subsection{Rapid Production of Minority Cultural Knowledge Benchmark Using RAG}

This study showcases the significant advantages of integrating RAG technology in generating and processing test datasets for minority and local cultural knowledge, addressing a critical limitation in traditional language model research that often overlooks these domains.

Previously, the reliance on fixed, mainstream-focused training datasets resulted in poor LLM performance when dealing with minority or local cultures. Constructing test datasets for these specialized areas was challenging due to data scarcity and often involved extensive time and resources.

Integrating RAG technology offers a transformative solution. By combining retrieval and generation capabilities, RAG allows models to dynamically access and utilize relevant information from diverse external sources, such as cultural dictionaries and local archives, during testing.

This approach offers several advantages:
\begin{enumerate}[noitemsep,leftmargin=*]
\item \textbf{Accelerated Test Dataset Generation}: RAG significantly reduces the time required to construct comprehensive test sets by rapidly retrieving vast amounts of relevant data.
\item \textbf{Enhanced Comprehensiveness}: By accessing diverse sources, RAG enables the creation of richer, more representative datasets that better reflect the complexities of minority and local cultures.
\item \textbf{Improved Quality and Accuracy}: RAG ensures that test questions are based on current, authoritative information, leading to more accurate evaluations even as cultural knowledge evolves.
\end{enumerate}
This RAG-based method represents a significant advancement in evaluating and optimizing LLM performance in multicultural contexts. It not only addresses the limitations of traditional approaches but also opens up exciting possibilities for future research by enabling the efficient and effective generation of high-quality test datasets for under-represented cultural domains.

\bibliographystyle{IEEEbib}
\bibliography{references}

\end{document}